\documentclass[fleqn,10pt]{wlscirep}
\usepackage[utf8]{inputenc}
\usepackage[english]{babel} 
\usepackage{setspace}
\usepackage{geometry}
\usepackage{graphicx}
\usepackage{hyperref}
\usepackage{booktabs}
\usepackage{amsmath}
\usepackage{caption}
\usepackage{subcaption}
\usepackage[utf8]{inputenc}
\usepackage[T1]{fontenc}

\title{Artefact-Aware Fungal Detection in Dermatophytosis: A Real-Time Transformer-Based Approach for KOH Microscopy}

\author[1,+]{Rana Gursoy}
\author[2,*,+]{Abdurrahim Yilmaz}
\author[3]{Baris Kizilyaprak}
\author[4]{Esmahan Caglar}
\author[2]{Burak Temelkuran}
\author[1]{Huseyin Uvet}
\author[3]{Ayse Esra Koku Aksu}
\author[5]{Gulsum Gencoglan}

\affil[1]{Department of Mechatronics Engineering, Yildiz Technical University}
\affil[2]{Division of Systems Medicine, Department of Metabolism, Digestion and Reproduction, Imperial College London}
\affil[3]{Department of Dermatology and Venereology, Istanbul Research and Training Hospital}
\affil[4]{Department of Bioengineering, Yildiz Technical University}
\affil[5]{Department of Dermatology and Venereology, Medicana Atakoy Hospital}

\affil[*]{Corresponding author: a.yilmaz23@imperial.ac.uk}
\affil[+]{These authors contributed equally to this work}

\begin{abstract}
Dermatophytosis is commonly assessed using potassium hydroxide (KOH) microscopy, yet accurate recognition of fungal hyphae is hindered by artefacts, heterogeneous keratin clearance, and notable inter-observer variability. This study presents a transformer-based detection framework using the RT-DETR model architecture to achieve precise, query-driven localization of fungal structures in high-resolution KOH images. A dataset of 2,540 routinely acquired microscopy images was manually annotated using a multi-class strategy to explicitly distinguish fungal elements from confounding artefacts. The model was trained with morphology-preserving augmentations to maintain the structural integrity of thin hyphae. Evaluation on an independent test set demonstrated robust object-level performance, with a recall of 0.9737, precision of 0.8043, and an AP@0.50 of 93.56\%. When aggregated for image-level diagnosis, the model achieved 100\% sensitivity and 98.8\% accuracy, correctly identifying all positive cases without missing a single diagnosis. Qualitative outputs confirmed the robust localization of low-contrast hyphae even in artefact-rich fields. These results highlight that an artificial intelligence (AI) system can serve as a highly reliable, automated screening tool, effectively bridging the gap between image-level analysis and clinical decision-making in dermatomycology.
\end{abstract}

\begin{document}

\flushbottom
\maketitle
\thispagestyle{empty}

\section*{Introduction}

Dermatophytosis is a common fungal infection of the nails and body. Early and accurate diagnosis is of critical clinical importance, as it directly influences treatment success \cite{tosti2019}. Conventional diagnostic methods primarily include direct microscopic examination with potassium hydroxide (KOH) and fungal culture, both of which remain widely used in clinical practice. However, these approaches are limited by prolonged analysis times, relatively low sensitivity, and significant inter-observer variability, particularly in microscopic evaluation \cite{gupta2020, levitt2010tinea}.

The KOH preparation, one of the most frequently employed method for microscopic confirmation, dissolves keratin and enhances the visibility of fungal hyphae and spores. However, direct KOH microscopy often presents diagnostic challenges due to residual keratin, air bubbles, debris, and artefactual structures that may resemble fungal elements, contributing to inter-observer variability and reduced sensitivity \cite{Velasquez2017,Lim2021}. Building a robust detection system requires addressing these inherent variabilities in samples prepared with KOH. In particular, artefacts that visually mimic fungal structures represent a major source of diagnostic ambiguity, not only for human observers but also for automated systems trained without explicit artefact-aware strategies. These artefacts also complicate automated analysis, particularly for object-detection models that rely on accurate spatial localization \cite{graham2019hovernet,bibin2017malaria}.

In recent years, artificial intelligence (AI) and deep learning methodologies have increasingly been adopted to address diagnostic challenges in medical image analysis, particularly in settings where subtle, low-contrast structures must be identified within complex visual backgrounds \cite{litjens2017}. Through their ability to learn complex representations from large datasets, these models can detect subtle and low-contrast patterns that are not readily discernible under conventional visual inspection \cite{shen2017}. Deep learning has achieved expert-level performance across multiple medical imaging domains, including radiology and digital pathology \cite{lakhani2017,rajpurkar2017}. In dermatology specifically, state-of-the-art models have demonstrated diagnostic accuracy comparable to, and in some cases exceeding, that of experienced dermatologists in the recognition of skin lesions \cite{esteva2017,brinker2019,ramesh2022}.

To place fungal hyphae detection in a broader context of microscopic image analysis, recent deep learning applications in related domains are briefly reviewed below. Microscopic image analysis has likewise become an important focus within medical AI, where deep learning models have been successfully applied to detect small, morphologically variable, and low-contrast biological structures \cite{zhang2024sperm,ponnusamy2020tb}. Strong performance has also been demonstrated in blood-smear microscopy, including malaria parasite detection using convolutional neural network (CNN) based or hybrid architectures \cite{raj2016malaria}. Convolutional and transformer-based models have further enabled reliable identification of parasitic forms and accurate detection of abnormal epithelial cells in cervical cytology \cite{bibin2017malaria,allehaibi2019cervical}. In addition, hybrid deep feature fusion strategies-such as those implemented in the DeepCervix framework-have advanced automated cervical cell analysis \cite{rahaman2021deepcervix}. Comparable progress in histopathology, including precise nuclei segmentation and classification, highlights the capacity of modern object-detection frameworks to localize fine, irregular, and artefact-obscured targets within complex visual fields \cite{graham2019hovernet,naylor2018regression}. These shared imaging characteristics (small object size, low contrast, complex textures, and the presence of distracting artefacts) closely mirror the challenges observed in KOH microscopy, making fungal hyphae a suitable and clinically relevant target for object-detection approaches.

Deep learning-based studies on dermatophytosis have proliferated rapidly in recent years. Yılmaz et al. (2022) used CNN-based models on KOH microscopy images and achieved 96.2\% accuracy in classifying infected versus healthy samples \cite{yilmaz2022}. Jansen et al. (2022) employed a U-Net-based segmentation model on histologic nail sections, identifying fungal structures with a Dice score of 91\% \cite{jansen2022}. Decroos et al. (2021) developed a deep learning model for histopathologic sections and obtained 92.5\% sensitivity, proving non-inferiority to conventional histopathologic diagnosis \cite{decroos2021}. Koo et al.\ applied a YOLOv4-based regional convolutional neural network to KOH microscopy images of superficial fungal infections, demonstrating that automated hyphae detection with bounding boxes is feasible in routine practice \cite{koo2021}. However, this study was limited to superficial mycoses rather than KOH preparations, relied on an earlier CNN architecture, and did not address the substantial artefactual heterogeneity characteristic of dermatophytosis microscopy. Consequently, despite increasing interest in automated analysis, most available research still focuses on image-level classification or pixel-wise segmentation, and object-level detection approaches in routine KOH microscopy have not yet been systematically investigated. Crucially, to date, the strategy of explicitly modeling artefactual structures as a distinct detection class to mitigate visual confusion has not been a primary focus of prior research, despite their well-recognized impact on diagnostic reliability in KOH microscopy \cite{garg2021mimickers}. This factor is particularly relevant for automated detection systems, where visually similar mimics may pose a challenge to model specificity, potentially affecting their seamless integration into clinical practice. 
 
From a clinical perspective, determining the precise location and extent of hyphae is essential for assessing fungal burden, highlighting the insufficiency of image-level classification. While visually similar artefacts in KOH preparations frequently confound conventional local feature-based detectors, transformer-based architectures leverage global self-attention to effectively discriminate between biological structures and mimics, as evidenced by recent success in similar microscopic tasks \cite{xu2024,chen2024,nakarmi2024}.  

In this study, we developed and evaluated an AI-based detection system for the automated and precise localization of fungal elements in routine KOH microscopy. All microscopic images were obtained using this routine preparation technique, thus capturing the heterogeneity, noise, and artefact patterns commonly encountered in real-world clinical practice, which are known to challenge the generalizability of AI-based systems \cite{litjens2017,shen2017}. Based on these considerations, we selected the RT-DETR (Real-Time Detection Transformer) model \cite{lv2023} as the object detection backbone. This study involved constructing an expert-verified and carefully curated dataset, where both fungal elements indicative of dermatophytosis and confounding artefactual structures were explicitly annotated as distinct classes. By training the RT-DETR model to simultaneously detect and distinguish these entities under standardized conditions, we assessed its performance using established detection metrics. Through this framework, we aimed to provide a fully automated and reproducible approach for the localization of fungal elements in routine KOH microscopy that remains robust against real-world artefactual interference.

\section*{Materials and Methods}

\begin{figure}[ht]
\centering
\resizebox{\textwidth}{!}{\includegraphics{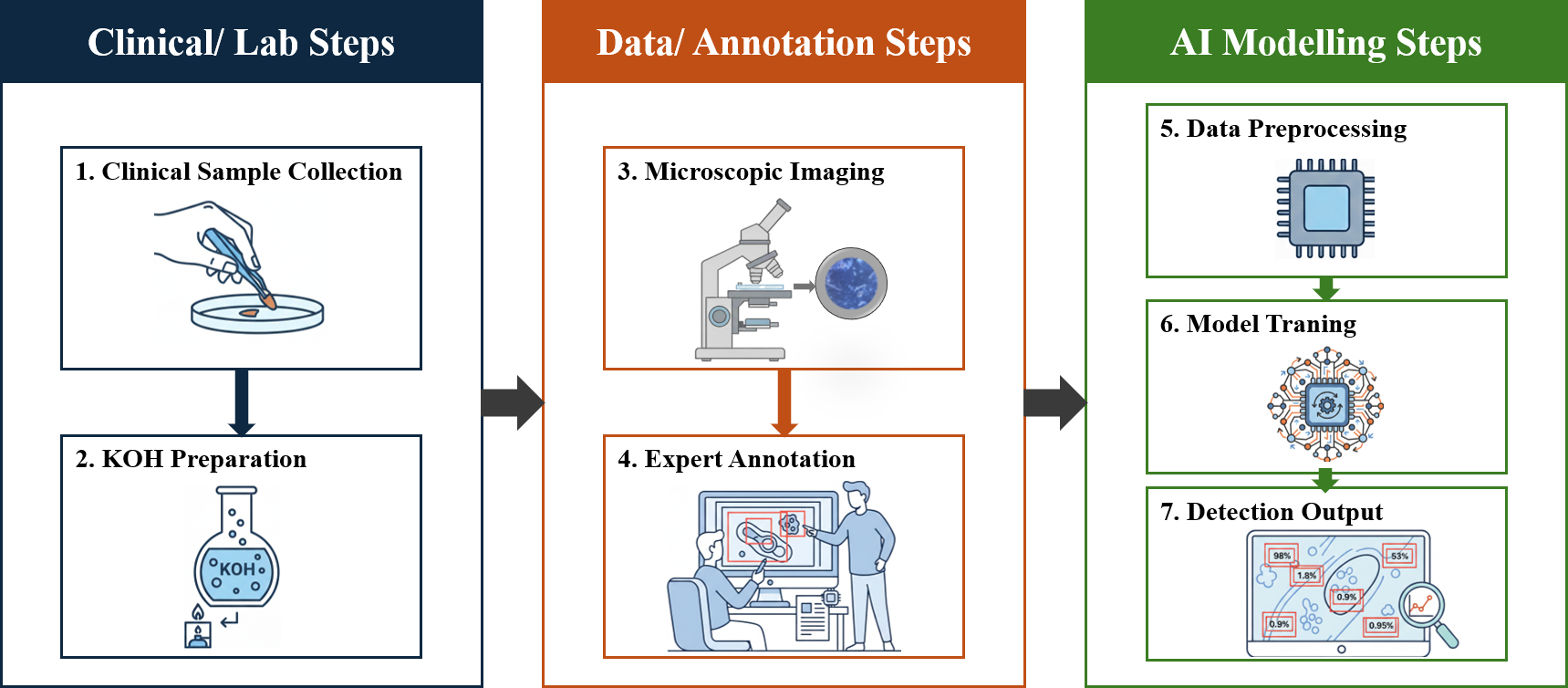}}
\caption{Schematic overview of the study workflow: clinical sample collection, KOH preparation, microscopic imaging, expert annotation, and artificial intelligence modelling steps.}
\label{fig:workflow}
\end{figure}

The complete study pipeline from clinical specimen collection and KOH preparation, through high-resolution microscopic imaging and expert bounding-box annotation, to object detection model training, validation, and inference is schematically illustrated in Figure~\ref{fig:workflow}.

\subsection*{1. Clinical Sample Collection and Microscopic Imaging}

The study involving human-derived specimens was approved by the Ethics Committee for Clinical Research under protocol number 2011-KAEK-50, with approval number 20. Human scrapings were collected at Istanbul Research and Training Hospital under sterile conditions from patients with clinically and/or mycologically confirmed dermatophytosis. The specimens were incubated in 10--20\% KOH solution to dissolve keratin, thereby enhancing the visibility of fungal hyphae and spores; nail specimens were incubated for 12 hours, whereas skin specimens were incubated for 1 hour. The cleared preparations were subsequently examined and imaged at high resolution using an light microscope.

\begin{table}[ht]
\centering
\caption{Descriptive statistics of the experimental dataset used for training and evaluation, highlighting the multi-class annotation strategy.}
\label{tab:dataset}
\begin{tabular}{llc}
\toprule
\textbf{Category} & \textbf{Component} & \textbf{Count / Value} \\
\midrule
Images & Total Microscopic Frames & 2,540 \\
Resolution & Pixel Dimensions & $2048 \times 2048$ \\
\midrule
Annotations & Fungal Elements (Positive Class) & 631 \\
 & Confounding Artefacts (Negative Class) & 381 \\
\bottomrule
\end{tabular}
\end{table}

A total of 2,540 microscopic images ($2048 \times 2048$ pixels) were acquired. All images were then reviewed and subjected to an initial quality-control assessment by two clinicians (Gulsum Gencoglan (G.G.) and Baris Kizilyaprak (B.K.)) who subsequently performed the manual annotation. The descriptive statistics and class distribution of the resulting dataset are summarized in Table~\ref{tab:dataset}.

\begin{figure}[ht]
\centering
\includegraphics[width=0.9\textwidth]{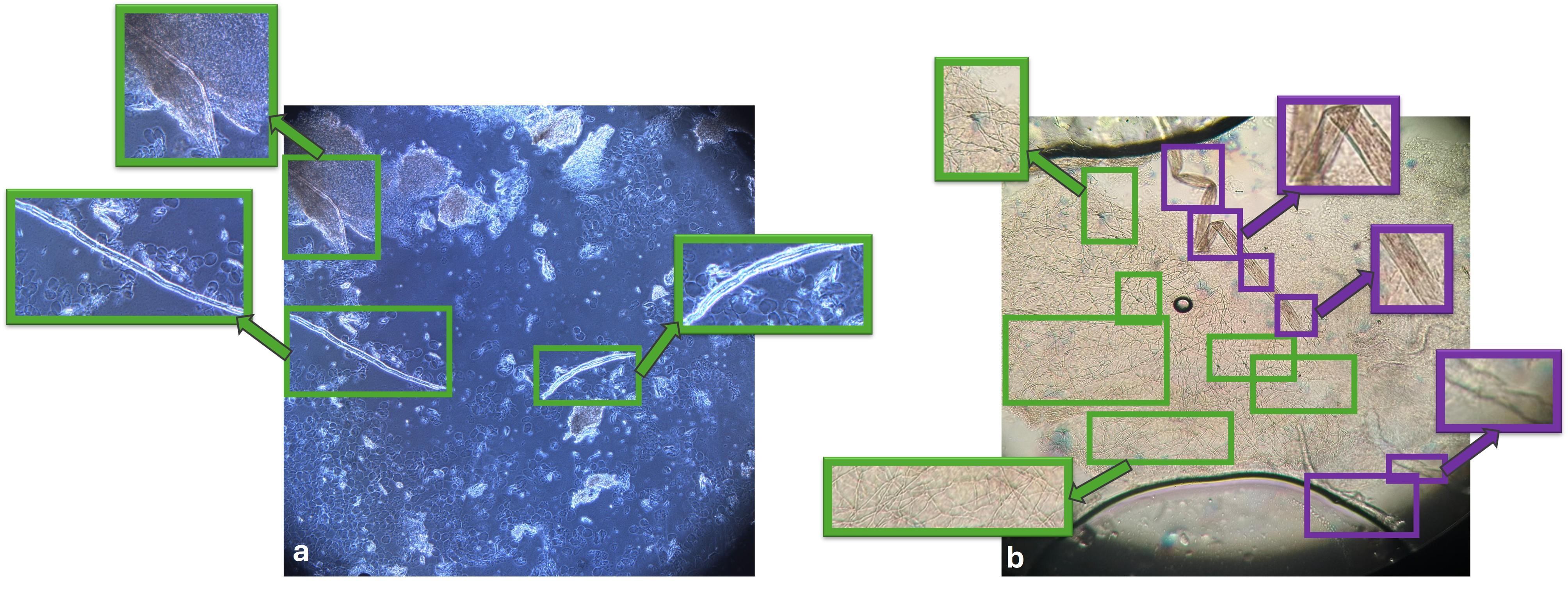}
\caption{(a,b) Representative KOH microscopy image illustrating the multi-class annotation strategy used in the dataset. Clear fungal hyphae are marked with green bounding boxes, while confounding artefactual structures (e.g., fibers, refractive edges) are highlighted with purple boxes. Insets provide magnified views of the annotated regions of interest.}
\label{fig:microscopic}
\end{figure}

The annotation process adopted a multi-class strategy to address visual ambiguity: specifically, 631 fungal elements (hyphae and spore clusters) were delineated as the primary target class, while 381 clinically relevant artefacts (such as keratin debris and fibers) were explicitly annotated as a distinct 'artefact' class. By deliberately labeling these mimics rather than treating them as background, the dataset was designed to teach the model to actively discriminate between true fungal structures and the diagnostic challenges frequently encountered in routine KOH microscopy. A representative example of these multi-class expert annotations is shown in Figure~\ref{fig:microscopic}.

\subsection*{2. Artificial Intelligence Model and Training Procedure}
\begin{figure}[ht]
    \centering
    \includegraphics[width=\linewidth]{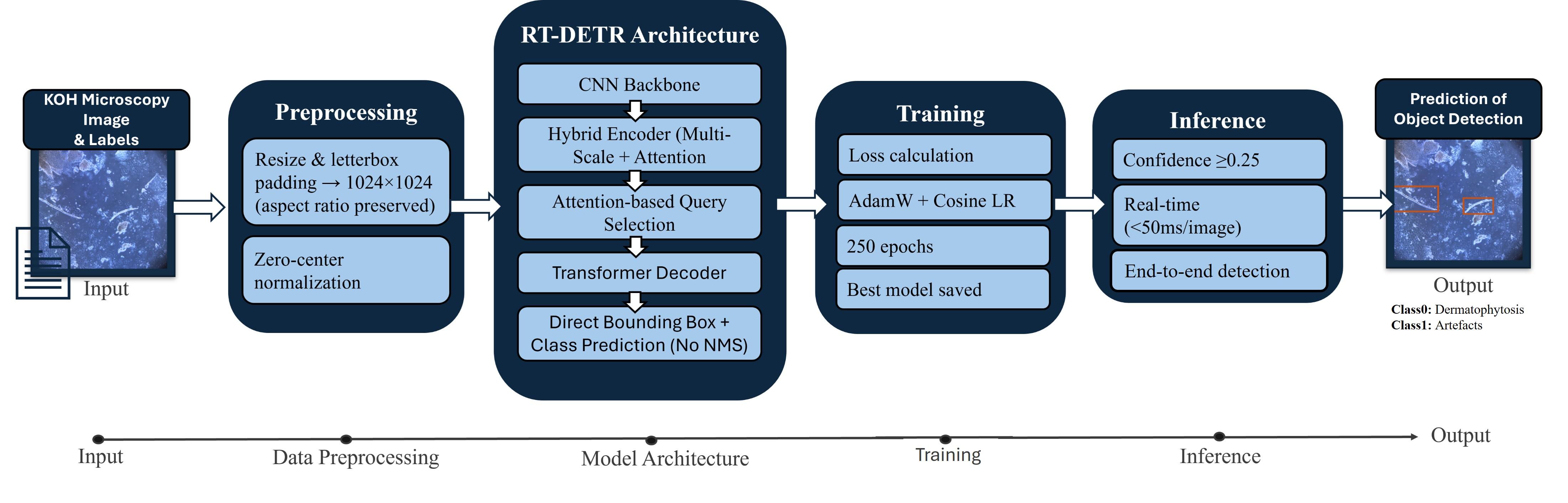}
    \caption{Overview of the model fungal detection pipeline. 
KOH microscopy images are preprocessed via letterbox resizing and normalization, 
then passed through the model architecture, which combines a CNN backbone, 
hybrid encoder, attention-based query selection, and a transformer decoder to 
produce bounding box predictions for both fungal elements and confounding artefacts. 
The model is trained using AdamW with cosine learning-rate scheduling, and inference is performed in real time with a confidence threshold of 0.25.}
    \label{fig:model-flow}
\end{figure}

Fungal element detection in KOH microscopy images was performed using the model, an end-to-end transformer-based object detector. The model employs an attention-driven query selection strategy within a hybrid CNN--Transformer encoder--decoder pipeline, enabling direct bounding box regression and classification. Combined with multi-scale feature fusion and efficient query refinement, the proposed architecture facilitates the detection of small, thin, and irregularly shaped fungal structures-such as hyphae and arthroconidia-while simultaneously distinguishing them from confounding artefactual mimics in dermatological microscopy. The overall model workflow, including preprocessing, architectural components, and inference logic, is illustrated in Figure~\ref{fig:model-flow}.

All experiments were conducted using the Ultralytics implementation of RT-DETR model (version~8.2.70), with the RT-DETR-L configuration pre-trained on the COCO dataset serving as the initialization. To ensure reproducibility, we note that training was performed on a workstation equipped with an NVIDIA RTX~4090 GPU (24~GB VRAM). The dataset was divided into training, validation, and test subsets following an \mbox{\(80\%\)--\(10\%\)--\(10\%\)} stratified split to preserve the distribution of both fungal and artefact instances. Prior to model ingestion, all images were resized to \(1024\times1024\) pixels using letterbox padding to maintain aspect ratio and were normalized using ImageNet statistics.

Model training followed the hyperparameters and optimization strategy defined in the custom training pipeline. Specifically, the model was trained for up to 250~epochs using the AdamW optimizer, an initial learning rate of \(5\times10^{-4}\), cosine-annealed decay with warm-up, a batch size of~8, and mixed-precision computation to reduce memory overhead. Loss functions were weighted following standard the model practice (box~=~7.5, classification~=~1.0), and early stopping with a patience of 50~epochs was applied to prevent overfitting. In line with the morphological sensitivity of fungal hyphae and the need to preserve distinct artefact features, aggressive augmentations such as MixUp were intentionally disabled. Standard geometric variations were utilized, including horizontal flipping (\(p=0.2\)), scaling (\(\pm20\%\)), translation (\(\pm5\%\)), and minor rotations (\(\pm2^\circ\)). These choices ensured preservation of clinically relevant fungal morphology while still providing sufficient variability for robust learning.

Model selection was based on validation performance measured by \(\text{mAP}_{0.5:0.95}\), which reflects detection consistency across different intersection-over-union thresholds for all classes. During inference, predictions were generated using a confidence threshold of~0.25. Due to its transformer-based matching mechanism, The model, enabling real-time prediction (\(\textless 50\,\text{ms}\) per image on RTX~4090) and improving reliability when fungal structures appear in dense or overlapping formations. This end-to-end inference pipeline is summarized in Figure~\ref{fig:model-flow}, which visualizes the progression from microscopy image preprocessing to final fungal region localization to final multi-class localization.

\section*{Results}

The performance of the detection framework was evaluated through an integrated hierarchy, progressing from localized object-level accuracy to patient-level diagnostic reliability. All quantitative metrics derived from the independent test set are summarized in Table~\ref{tab:integrated_metrics}, providing a unified view of the model's technical and clinical efficacy.

\begin{table}[ht]
\centering
\caption{Comprehensive performance evaluation of the model, encompassing both object-level detection and image-level diagnostic classification.}
\label{tab:integrated_metrics}
\begin{tabular}{lll}
\toprule
\textbf{Evaluation Level} & \textbf{Metric} & \textbf{Value} \\
\midrule
\textbf{Object-Level} & Precision & 0.8043 \\
\textit{(Detection Performance)} & Recall (Sensitivity) & 0.9737 \\
 & F1-Score & 0.8810 \\
 & AP@0.50 (\%) & 93.56 \\
 & AP@0.50:0.95 (\%) & 78.40 \\
 & Mean IoU & 0.8560 \\
\midrule
\textbf{Image-Level} & Accuracy & 0.9882 \\
\textit{(Clinical Diagnosis)} & Sensitivity (Recall) & 1.0000 \\
 & Specificity & 0.9818 \\
 & Precision & 0.9674 \\
 & F1-Score & 0.9835 \\
 & Missed Diagnoses (FN) & 0 \\
\bottomrule
\end{tabular}
\end{table}

At the object level, the framework achieved a recall of 0.9737 and a precision of 0.8043, effectively identifying the vast majority of fungal elements while maintaining a balanced false-positive rate. The model demonstrated high precision in localizing complex fungal structures within infected fields, supported by an AP@0.50 of 93.56\% and a mean IoU of 0.8560. These metrics reflect strong spatial agreement between predicted bounding boxes and expert annotations. The transformer-based architecture demonstrated stable performance in identifying low-contrast fungal hyphae, including in artefact-rich fields. The attention-driven query selection mechanism enabled separation of overlapping hyphae in densely populated microscopic fields.

\begin{figure}[ht]
\centering
\includegraphics[width=0.8\textwidth]{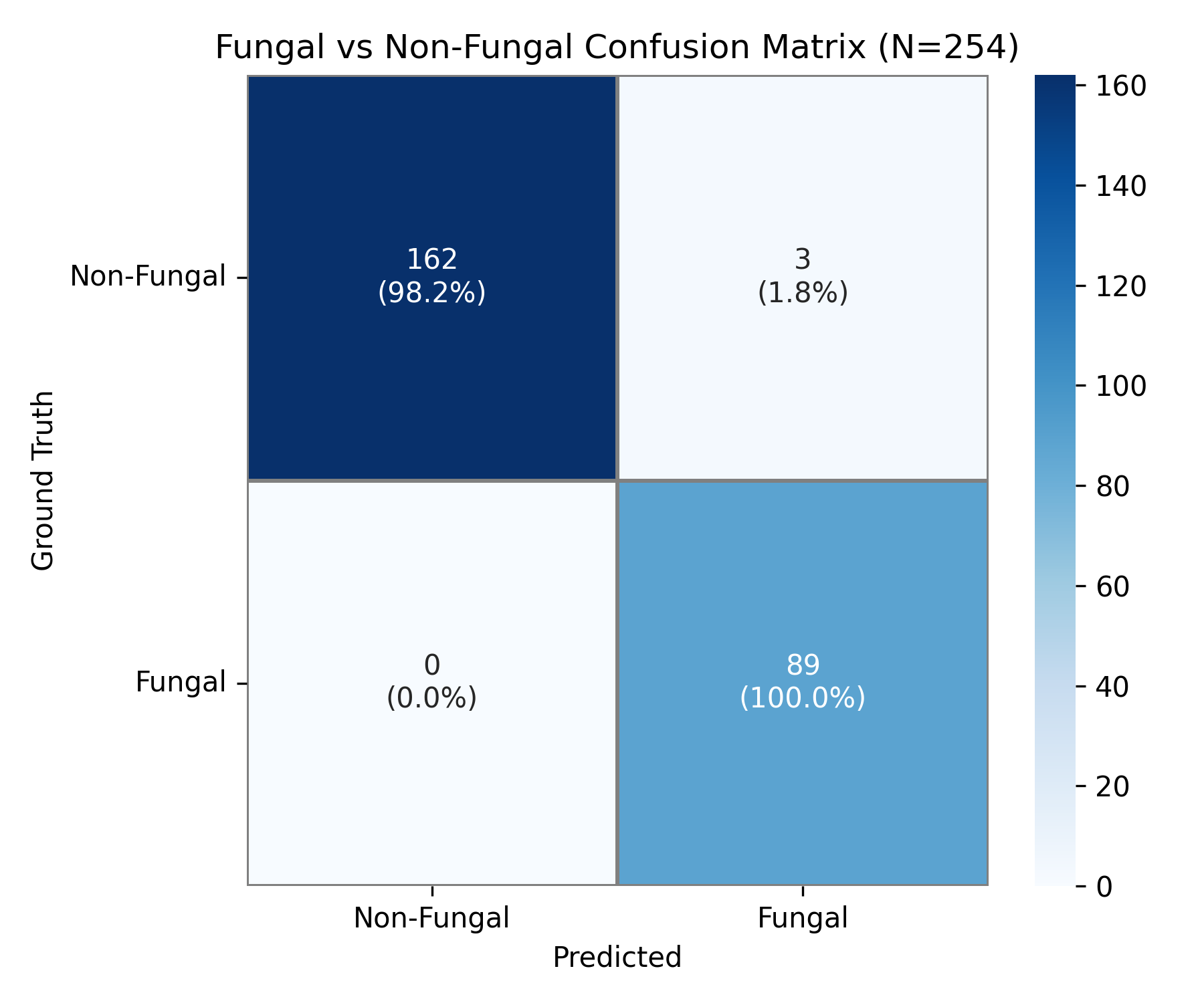}
\caption{Image-level confusion matrix evaluating diagnostic performance across 254 test samples (89 positive, 165 negative). The model correctly identified all infected images with zero false negatives, yielding 98.8\% overall accuracy.}
\label{fig:confusion_matrix}
\end{figure}

The primary clinical utility of the system was assessed by aggregating localized detections for image-level diagnosis across the entire independent cohort ($n=254$ images obtained from individual clinical cases). Following the clinical principle that a single confirmed hypha is sufficient for diagnosis, a case was classified as positive if at least one fungal element was detected with high confidence ($>0.25$). Under this strategy, the model achieved a sensitivity of 100\%, correctly identifying all 89 positive cases without missing any fungal infections ($FN=0$). The system demonstrated robust performance across the full distribution of samples, achieving a specificity of 98.18\% and an overall diagnostic accuracy of 98.82\%. Figure~\ref{fig:confusion_matrix} visualizes these diagnostic outcomes in a confusion matrix, confirming the model's reliability as an automated screening tool that prioritizes clinical safety.

\section*{Discussion}

This study presents an object detection model for localizing fungal structures in KOH microscopy images of dermatophytosis. Previous deep learning work in this field has largely focused on image-level classification of infected versus non-infected samples or pixel-wise segmentation of fungal elements \cite{yilmaz2022,jansen2022,decroos2021}. In contrast, our approach targets explicit object-level localisation with bounding boxes, utilizing a multi-class detection strategy to distinguish true fungal elements from confounding artefacts, thereby aiming to provide interpretable and explainable visual cues on routine microscopy images.

Modern detection architectures have recently been applied to the automated analysis of fungal infections. Koo et al.\ employed a YOLOv4-based one-stage convolutional detector for hyphae detection in KOH microscopy images, demonstrating that automated bounding-box localisation is feasible in clinical practice \cite{koo2021}. While their work provided an important proof of concept for CNN-based object detection, YOLOv4 relies on convolutional feature hierarchies and predefined anchor mechanisms, which represent a different design paradigm compared to transformer-based detectors. Moreover, their study focused on superficial mycoses under relatively controlled imaging conditions, with less visual complexity than is typically encountered in KOH preparations. In contrast, the present study evaluates RT-DETR model, a transformer-based detector that performs end-to-end object detection, using attention-driven query selection and global contextual reasoning. This architecture was evaluated in high-resolution KOH microscopy settings, where fungal hyphae appear small, low-contrast, and are often embedded within dense backgrounds containing keratin debris and preparation-related artefacts. By explicitly modelling mimics-such as fibres and air bubbles-as a distinct class during training, our model leverages global contextual reasoning to actively suppress false positives, a capability often limited in local-feature-dependent CNNs. From a computational perspective, the model enables stable real-time inference in our experimental setting (below 50 ms per image), while maintaining a simplified detection pipeline without post-processing. These characteristics support not only efficient image-level analysis but also scalability to large microscopic fields, which is critical for real-time or semi-automated slide examination. Taken together, while YOLOv4-based approaches have demonstrated the feasibility of automated hyphae detection, alternative detection frameworks that emphasize global contextual information and end-to-end prediction may offer complementary advantages in artefact-rich KOH microscopy settings.

Object-level localisation is also closely aligned with the way clinicians read KOH slides. In daily practice, fungal elements must be identified among numerous mimickers, including keratin debris, air bubbles, fibres and staining artefacts, many of which can resemble hyphae and contribute to diagnostic variability \cite{Velasquez2017}. Less-experienced observers in particular may confuse hyphae with these structures, a well-documented limitation of KOH microscopy that contributes to diagnostic variability \cite{Lim2021}. Moreover, routine KOH examination often requires manual scanning of multiple large microscopic fields, which is time-consuming and may further increase reader fatigue and variability \cite{litjens2017}. By drawing bounding boxes around regions with a high probability of containing fungal elements while ignoring learned artefacts, the proposed model can highlight diagnostically relevant regions, thereby improving interpretability and transparency in the decision-making process. Such visual localization aligns with principles of explainable artificial intelligence, which emphasise transparency and human interpretability as key requirements for clinical deployment of AI systems \cite{tjoa2020survey}. In addition, the number and distribution of detected regions could in future be used to approximate fungal burden or to monitor changes during treatment.

Quantitative analysis of the misclassified instances revealed nine false positive and one false negative predictions. The false positives were primarily associated with synthetic fibers, scratch marks, overlapping cell borders, and sharp keratin edges that visually mimicked hyphal morphology, whereas the single missed detection corresponded to an extremely faint or out-of-focus hyphal fragment exhibiting very low contrast against a dense keratin background. These findings represent the technical edge cases encountered during the detection of subtle biological structures in heterogeneous KOH preparations. Overall, the quantitative results support the feasibility of this approach. On the independent test set, the model achieved a recall of 0.9737 and a precision of 0.8043, yielding an F1-score of 0.8810, indicating that the system identifies the vast majority of annotated fungal elements with a limited but non-negligible number of false positives. Importantly, when object-level predictions were aggregated for image-level diagnosis, the system achieved 100\% sensitivity and 98.8\% accuracy, correctly identifying all positive cases (n=89) without missing any fungal infections. This perfect image-level sensitivity confirms that even if an isolated hyphal fragment is missed, the model captures sufficient visual evidence to correctly flag the pathology. Localisation performance was also strong, with AP@0.50 of 93.56\%, AP@0.50:0.95 of 78.40\%, and a mean IoU of 0.8560. Evaluating AP across multiple IoU thresholds provides a more stringent measure of spatial agreement than a single-threshold metric and is consistent with current practice in object detection. The performance levels observed here are comparable to those reported for other microscopic detection tasks, including leukocyte detection, parasitic egg localisation, and hyphae detection in KOH images~\cite{koo2021,xu2024,chen2024,nakarmi2024}.

\begin{figure}[ht]
\centering
\includegraphics[width=\textwidth]{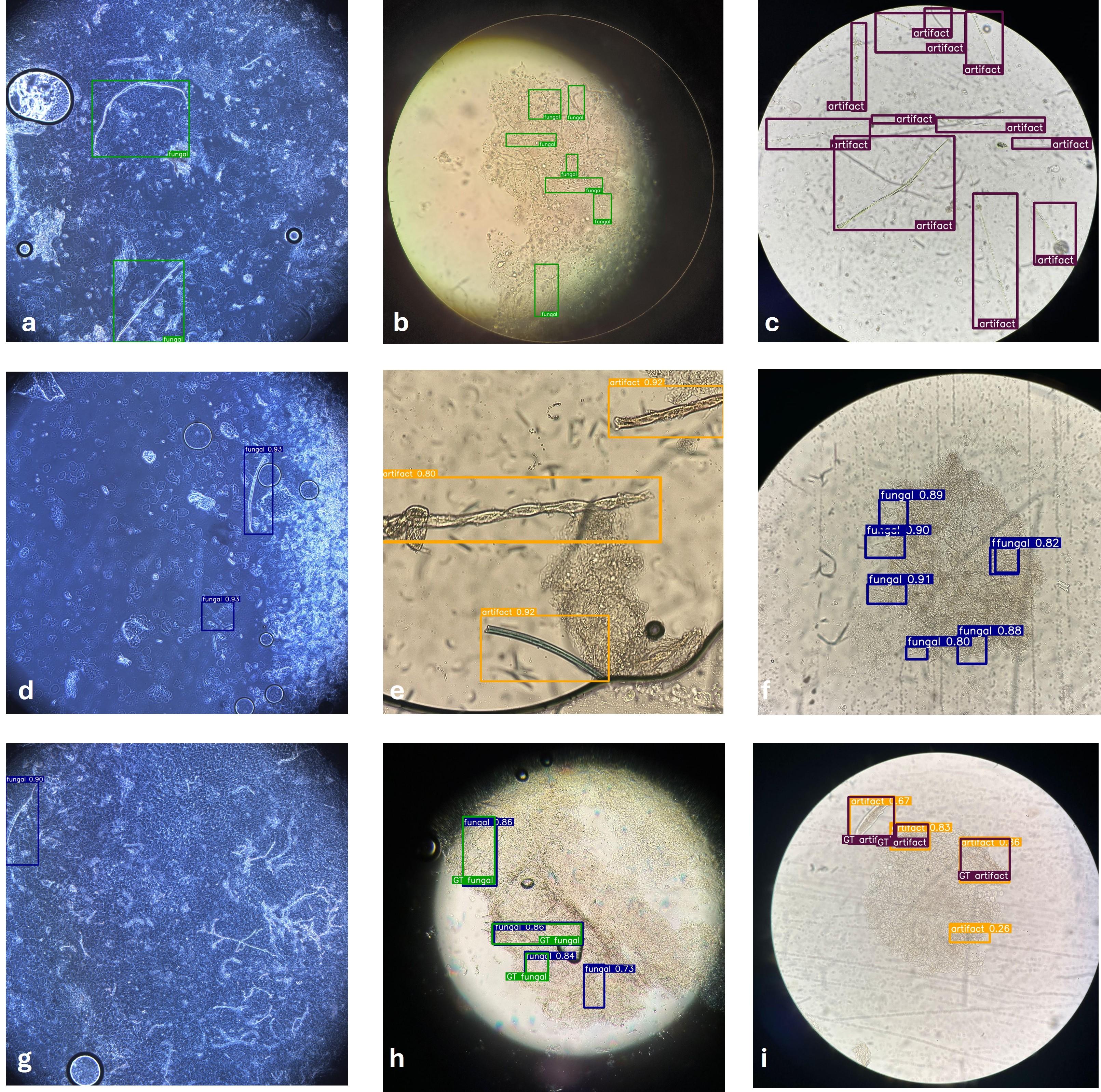}
\caption{Comprehensive qualitative evaluation and failure analysis of the proposed model.
(a--c) Expert-defined Ground Truth (GT) annotations illustrating the primary training categories: fungal elements (fungi; green bounding boxes) and artefacts (purple bounding boxes).
(d--f) Model inference results showing predicted fungal elements with corresponding confidence scores; panel (e) highlights successful discrimination between true fungal structures and visually similar artefactual mimics.
(g--i) Comparative overlays of GT annotations and model predictions, including fungal detections (blue bounding boxes) and artefact predictions (orange bounding boxes), enabling direct visual assessment of class-wise localization performance.
The high spatial overlap and correct class assignment across heterogeneous backgrounds demonstrate the model's robustness against common KOH microscopy artefacts.}
\label{fig:qualitative_analysis}
\end{figure}

To qualitatively evaluate the model’s performance, a comparative failure analysis was conducted across different detection scenarios in Figure~\ref{fig:qualitative_analysis}. The top row, comprising Panels~\ref{fig:qualitative_analysis}a, \ref{fig:qualitative_analysis}b, and \ref{fig:qualitative_analysis}c, illustrates the expert-verified ground truth (GT) annotations, highlighting the high density of fungal structures. The middle row shows the model's independent inference; while Panels~\ref{fig:qualitative_analysis}d and \ref{fig:qualitative_analysis}f demonstrate high-confidence detections, Panel~\ref{fig:qualitative_analysis}e identifies "mimickers", non-fungal artefacts that visually resemble hyphae, and labels them as artefacts to prevent diagnostic confusion.

The bottom row (Panels~\ref{fig:qualitative_analysis}g, \ref{fig:qualitative_analysis}h, and \ref{fig:qualitative_analysis}i) specifically illustrates instances of misclassification and the limitations of the detection framework. In Panel~\ref{fig:qualitative_analysis}g, the model exhibits a false positive error by identifying a non-fungal structure as a hypha, which would lead to a healthy sample being incorrectly flagged as infected (dermatophytosis). Similarly, Panel~\ref{fig:qualitative_analysis}h reveals a critical misclassification where the model identifies an ovoid, non-fungal structure as dermatophytosis, further demonstrating the system's tendency to over-predict in the presence of ambiguous biological morphology. Finally, Panel~\ref{fig:qualitative_analysis}i confirms that certain synthetic fibers and sharp keratin edges continue to trigger false detections despite the multi-class training strategy. These instances represent technical bottlenecks where the high morphological similarity between debris and true hyphae limits the model's precision in complex KOH preparations.

Several characteristics of the dataset and training strategy have likely contributed to these results. The images were acquired under routine conditions and deliberately included a wide range of KOH dissolution stages, from partially cleared preparations with abundant keratin to more transparent fields. This heterogeneity introduces considerable noise and artefacts, which makes the task more challenging for the model but also increases its relevance to real-world practice, where slides are rarely ideal. The use of high-resolution inputs and a transformer-based backbone allowed the network to combine local shape cues with broader contextual information from the surrounding tissue. This combination of fine-grained feature extraction and contextual modelling has also been shown to improve the detection of small, low-contrast structures in other microscopic imaging tasks \cite{graham2019hovernet}.

This study acknowledges limitations inherent to its single-centre design, necessitating external multi-centre validation to confirm generalisability. Regarding class definitions, while the current model employs a multi-class strategy to distinguish fungal elements from artefacts, it treats all fungal species as a single category. Future extensions should therefore incorporate richer annotation schemes to differentiate specific fungal morphologies. A critical clinical consideration is the differential diagnosis of dermatophytosis. As emphasized by recent studies, nail unit malignancies-specifically subungual melanoma and squamous cell carcinoma-frequently mimic the clinical presentation of dermatophytosis through features like hyperkeratosis and dystrophic discoloration \cite{curtis2025diagnosis,venturi2025squamous}. Since the proposed model is trained to detect fungal presence but not to exclude these mimickers, relying solely on automated screening could obscure concurrent malignancies. Therefore, future work aims to integrate anomaly detection mechanisms to flag atypical patterns for expert review. Finally, prospective reader studies will be essential to translate these image-level performance gains into tangible improvements in clinical diagnostic accuracy and reporting efficiency.

In conclusion, this study demonstrates that a transformer-based detection framework, reinforced by a multi-class training strategy, can reliably identify fungal elements even within the complex, artefact-laden environment of routine KOH microscopy. Crucially, the model’s perfect sensitivity at the image level suggests it can serve as a safe and effective screening tool, ensuring that no positive cases are overlooked. By translating abstract neural network predictions into interpretable visual cues, this approach bridges the gap between 'black-box' AI and clinical explainability. Ultimately, this localisation capability may facilitate integration into clinical workflows as a decision-support tool, potentially contributing to more consistent and standardized diagnostics in dermatomycology.

\section*{Data Availability Statement}

The datasets generated and analyzed during the current study are available from the corresponding author upon reasonable request. Raw microscopic images are not publicly available due to ethical and privacy considerations associated with human-derived biological material.

\newpage
\bibliography{sample}

\end{document}